\documentclass{article}
\usepackage[accepted]{tmlr}   
\usepackage{graphicx}
\usepackage{amsmath}
\usepackage{amssymb}
\usepackage{booktabs}
\usepackage{float}
\usepackage{multirow}
\usepackage{pifont}
\usepackage{algorithm}
\usepackage{algpseudocode}
\usepackage{xcolor}
\usepackage{xspace}
\usepackage[hidelinks]{hyperref}

\newcommand{\method}{\textsc{Forge}\xspace}
\newcommand{\cmark}{\textcolor{green!45!black}{\ding{51}}}  
\newcommand{\xmark}{\textcolor{gray!75}{\ding{55}}}        

\title{FORGE: Forward-Only Test-Time Adaptation for\\ Integer-Only Vision Models on Microcontrollers}

\author{%
Muhammad Rehan\\
SEECS, NUST, Islamabad, Pakistan\\
\texttt{mrehan.msee21seecs@seecs.edu.pk}
\AND
Haider Ali\\
FAST-NUCES, Islamabad, Pakistan\\
\texttt{i171095@nu.edu.pk}
\AND
Muhammad Ali Munir\\
SEECS, NUST, Islamabad, Pakistan\\
\texttt{mmunir.msee21seecs@seecs.edu.pk}
\AND
Moaz Amjad\\
SEECS, NUST, Islamabad, Pakistan\\
\texttt{mamjad.msee21seecs@seecs.edu.pk}
}

\def\month{08}
\def\year{2026}
\def\openreview{\url{https://openreview.net/forum?id=A45I5p25dd}}

\begin{document}
\maketitle

\begin{abstract}
Vision models deployed on microcontrollers (MCUs) are quantized to integer-only
arithmetic and run in inference-only runtimes that do not carry the machinery
backpropagation needs: the standard tool for adapting a model to the distribution
shift (sensor noise, blur, lighting) it meets in the field. Existing forward-only test-time adaptation (TTA) methods either run only on
server- or edge-GPU-class models (not true microcontroller integer execution), or require the
batch-normalization (BN) layers that integer deployment fuses away. We present a
forward-only TTA method that operates on \emph{deployed}, BN-folded, integer-only
convolutional networks. The key observation is that fusing BN into the preceding
convolution, a mandatory step for integer inference, destroys the statistics that
normalization-based adaptation relies on. We restore adaptation by re-normalizing each
folded convolution's per-channel output to its clean training statistics, using only
forward-pass estimates. The method (i) recovers most of gradient-based TENT's
accuracy gain ($+20.9$ vs.\ $+24.9$ points) and matches forward-only BN adaptation,
while being the only method that runs on a folded integer-only model; (ii) needs to
adapt only $3$ of $21$ layers (selected without seeing the test
corruptions) to recover $93\%$ of the benefit; (iii) survives single-sample streaming
with a batch-size-scaled momentum; and (iv) generalizes across three datasets (up to
200 classes) and two architectures. We validate bit-exact int8 convolution execution and deploy on an ESP32-S3,
where, measured with a Nordic PPK2 power profiler, the forward-only adaptation (a lightweight
fp32 recalibration around the int8 convolutions) costs only
\textbf{8.3\,mJ (6.8\% of inference energy)} and \textbf{21.9\,ms} on the deployed
SIMD-optimized model: forward-only adaptation is cheap on a real microcontroller.
\end{abstract}

\section{Introduction}
\begin{figure}[t]\centering
\includegraphics[width=\linewidth]{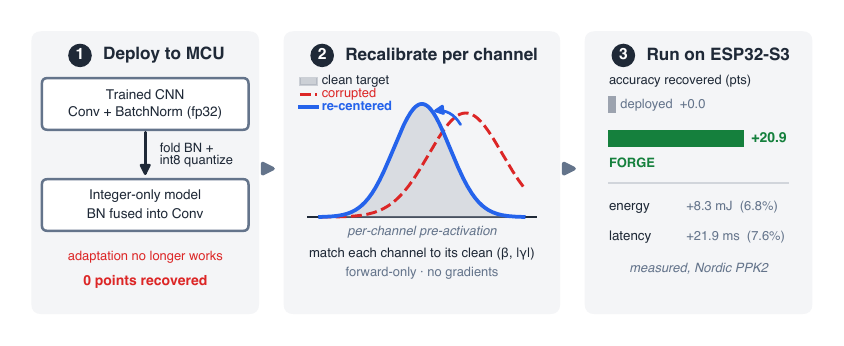}
\caption{\textbf{\method in one view.}
\textbf{(1)}~Deploying a CNN to a microcontroller folds batch normalization into the
preceding convolutions and quantizes to int8; the normalization statistics that
test-time adaptation relies on are baked in, so adaptation out of the box recovers
nothing ($+0.0$). \textbf{(2)}~\method restores adaptation \emph{forward-only}: under
corruption each folded channel's output distribution drifts (red), and \method re-centers
it onto its clean training target $(\beta,\,|\gamma|)$ (blue) from running per-channel
statistics, using no gradients. \textbf{(3)}~On an ESP32-S3 this recovers $+20.9$ accuracy
points at a measured $6.8\%$ extra energy ($8.3$\,mJ) and $21.9$\,ms on ESP-NN SIMD
kernels, profiled with a Nordic PPK2.}
\label{fig:teaser}
\end{figure}

On-device vision on microcontrollers is attractive for privacy, latency, and energy:
no data leaves the device and no network is required. To fit kilobytes of SRAM and
run on integer-only arithmetic units, models are quantized to int8 and their batch
normalization (BN) layers are \emph{folded} into the preceding convolutions. This is
mandatory for efficient integer inference (TFLite-Micro, CMSIS-NN, ESP-NN), but it has
a consequence that is rarely discussed: the deployed model can no longer be adapted.

Models degrade under distribution shift (noise, blur, weather, compression) that they
never saw during training~\citep{cifarc2019}. On a server, one simply fine-tunes or runs
test-time adaptation (TTA). On the deployed MCU model this is out of reach for two
compounding reasons.
First, the deployed inference-only runtime does not carry the machinery
backpropagation needs (an autograd graph, fp32 master weights, optimizer state), so
gradient-based TTA such as TENT~\citep{tent2021} or CoTTA~\citep{cotta2022} does not run
on it as deployed; on-device training frameworks \emph{can} backpropagate through
quantized models~\citep{lin2022odt,deutel2024cortexm,buron2025qpu}, but only by
reintroducing that machinery at substantial memory and compute cost (Sec.~\ref{sec:related}).
Second, the forward-only
alternative, recalibrating BN statistics~\citep{bnadapt2020,leantta2025}, requires BN
layers, which folding has removed. The deployed model thus degrades exactly as much as
its float counterpart, but \emph{none} of the existing remedies apply to it.

Recent forward-only TTA methods do not close this gap. FOA~\citep{foa2024},
ZOA~\citep{zoa2025}, and PACE~\citep{pace2026} are gradient-free but are demonstrated on
server- and edge-GPU-class hardware, not under true microcontroller integer execution;
PEA~\citep{pea2026} is gradient-free but does not target quantization; and
TinyTTA~\citep{tinytta2024} runs on an MCU but is gradient-based (it only reduces
backprop memory). None reports the one quantity that matters for a battery device:
measured on-device energy.

We close this gap. Our contributions are:
\begin{itemize}\itemsep2pt
\item We identify and empirically demonstrate the \emph{adaptation gap}: folding BN for
integer deployment removes the statistics that forward-only normalization adaptation
needs, eliminating its $+20$-point recovery (Sec.~\ref{sec:gap}).
\item We propose \method (Forward-Only Recalibration for the ed\textsc{ge}), a
forward-only per-channel recalibration that restores
adaptation on the deployed folded integer-only model, recovering most of gradient-based
TENT's gain and matching forward-only BN adaptation, while being uniquely deployable
(Sec.~\ref{sec:method}, Table~\ref{tab:baselines}).
\item We show adaptation is needed at only a few layers: $3$ of $21$ layers, selected on
held-out corruptions, recover $93\%$ of the benefit, and adapting the right subset
\emph{exceeds} adapting all layers (Sec.~\ref{sec:selective}).
\item We characterize single-sample streaming and show a batch-size-scaled momentum
rescues it; and we demonstrate generalization across three datasets (CIFAR-10, CIFAR-100,
and Tiny-ImageNet, spanning up to 200 classes) and two architectures.
\item We validate bit-exact int8 convolution execution and deploy on an ESP32-S3, reporting the
first \emph{measured} energy cost of forward-only TTA on an MCU: $8.3$\,mJ ($6.8\%$ of
inference energy), $21.9$\,ms, on ESP-NN SIMD kernels (Sec.~\ref{sec:device}). The
adaptation is mixed-precision by design (int8 convolutions with a lightweight fp32
recalibration); we report its data types and costs explicitly (Sec.~\ref{sec:device}).
\end{itemize}

A central message is that adapting a deployed integer-only model does not require a
complex algorithm; it requires recognizing what folding removes and restoring it
cheaply. The adaptation step itself is intentionally minimal: no gradients, no
learnable parameters, and no extra forward passes. That minimalism is precisely what
makes it run, at negligible energy, inside a deployed runtime that does not backpropagate. The novelty is in the problem
(the deployment gap), the efficiency (only a few layers, single-sample streaming), and
the evidence (bit-exact int8 execution on real hardware with measured energy), not in the
adaptation rule.

\section{Related Work}
\label{sec:related}
\paragraph{Gradient-based TTA.} A large family of test-time adaptation methods updates
the model by backpropagation: TENT~\citep{tent2021} minimizes prediction entropy over the
BN affine parameters; SHOT~\citep{shot2020} uses information maximization and pseudo-labels;
EATA~\citep{eata2022} and SAR~\citep{sar2023} improve efficiency and stability under
non-i.i.d.\ streams; MEMO~\citep{memo2022} enforces augmentation consistency; and
CoTTA~\citep{cotta2022} and NOTE~\citep{note2022} target the continual setting. These methods
define the accuracy ceiling, but the obstacle to deploying them on an MCU is structural
rather than a matter of efficiency: backpropagation needs the stored activation graph and
optimizer state of a floating-point model, neither of which an integer-only runtime
(TFLite-Micro, ESP-NN) materializes. Even memory-reduced variants still differentiate, so
they remain off the integer path. We use TENT and CoTTA as (undeployable) accuracy
references.

\paragraph{Forward-only normalization adaptation.} Adapting normalization statistics
forward-only is a long-standing idea: AdaBN~\citep{adabn2018} replaces source BN statistics
with target-domain ones, and \citet{hecheng2018} recalibrate BN statistics from limited
unlabeled data to restore accuracy after quantization and pruning. More recently,
prediction-time BN adaptation~\citep{bnadapt2020,predbn2020} and LeanTTA~\citep{leantta2025}
recalibrate normalization statistics at test time without gradients, and
T3A~\citep{t3a2021} adjusts the classifier with a bank of stored prototypes. They are the
closest forward-only baselines, and they share a dependence \method removes: \emph{all of
them assume a live batch-normalization layer to refresh} (AdaBN, He \& Cheng, BN-adapt,
LeanTTA) or a cached representation of the source feature space (T3A). This is exactly the
regime integer deployment destroys: once BN is folded into the convolution for integer-only
inference the statistics are gone, and the feature bank is not shipped, so none of these
methods has anything to update. \method differs precisely by operating \emph{after} the
fold: it carries only two per-channel constants per former-BN site, recorded for free at
fold time, and recovers the statistics post-fold, which is what lets it run on the deployed
model. The mechanism is deliberately close to this prior line of work; the contribution is
making it run where the BN layer no longer exists (Sec.~\ref{sec:limits}).

\paragraph{Forward-only TTA for quantized models.} FOA~\citep{foa2024} and
PACE~\citep{pace2026} adapt quantized ViTs with derivative-free optimization (CMA-ES);
ZOA~\citep{zoa2025} uses zeroth-order estimation and covers CNNs. All three run on GPU or
edge-GPU hardware (PACE's only on-device number is a Jetson Xavier NX throughput), and
none reports MCU energy. PEA~\citep{pea2026} is
forward-only and architecture-agnostic but does not address quantization.

\paragraph{TTA on microcontrollers.} TinyTTA~\citep{tinytta2024} is the closest prior
work and the only other TTA method validated on an MCU (STM32H747), but it is
gradient-based: a self-ensemble + early-exit strategy that \emph{reduces} backprop memory
rather than eliminating gradients, and it adapts the network while its normalization
layers are still present. On the deployed \emph{folded} integer-only model it therefore
falls in the same bucket as TENT and BN-adapt (Table~\ref{tab:position}): it needs the
gradients and the live normalization layers that folding removes, so it does not run on
the deployed model as-is and a like-for-like accuracy row would not be apples-to-apples.
\method differs on every axis that matters for deployment: forward-only, true
integer-only, on a folded model, with measured energy. The comparison below makes this
quantitative: the two methods solve the same problem (small-batch TTA on microcontrollers)
from opposite ends, TinyTTA keeping gradients (and adding trained early-exit heads) before
the fold, \method forward-only on the deployed folded int8 model. Entries are from the
respective papers (\citealp{tinytta2024} for TinyTTA, this work for \method).

\par\medskip
\noindent\begin{minipage}{\linewidth}
\centering
\footnotesize
\textbf{\method vs.\ TinyTTA, the closest prior MCU TTA method.}\\[3pt]
\setlength{\tabcolsep}{4pt}
\begin{tabular}{lll}\toprule
property & TinyTTA~\citep{tinytta2024} & \method (ours) \\\midrule
adaptation mechanism & gradient-based early-exit ensemble & forward-only recalibration \\
gradients / optimizer state & yes (reduced via early exit) & none \\
normalization layers & trained early-exit heads, pre-fold & post-fold $(\beta,|\gamma|)$ \\
runs on deployed folded int8 model & no & yes \\
batch size $1$ & yes & yes (window-matched momentum) \\
adaptation memory & early-exit heads + backprop buffers & $\approx$6\,KB SRAM + 6\,KB Flash (fp32 stats) \\
measured on-device energy & not reported & $8.3$\,mJ ($6.8\%$) \\
hardware validated & STM32H747 (Cortex-M7) & ESP32-S3 \\\bottomrule
\end{tabular}
\end{minipage}\par\medskip

\paragraph{On-device training of quantized models.} A separate line of work
backpropagates through quantized networks \emph{on the device}: MCUNet's tiny training
engine fits training into 256\,KB via sparse layer updates and quantization-aware
scaling~\citep{lin2022odt}, and others train fully quantized models on Cortex-M
MCUs~\citep{deutel2024cortexm} or cut training memory with quantized parameter
updates~\citep{buron2025qpu}. These systems show that on-device backpropagation through
quantized models is feasible, but they pay for it: they reintroduce the gradient machinery
(an autograd graph, higher-precision master weights, optimizer state, and quantized-gradient
bookkeeping) that a pure inference runtime omits, and they target on-device \emph{training}
rather than the inference-only deployment \method adapts within. \method occupies the
opposite corner: it adds no gradient machinery and runs inside the deployed int8 inference
engine, trading the generality of gradient updates for negligible cost.

\paragraph{Quantization and on-device inference.} Integer-only
inference~\citep{jacob2018} folds batch normalization~\citep{batchnorm2015} into the
preceding convolution and quantizes weights and activations to int8~\citep{krishnamoorthi2018,
nagel2021,lsq2020}. This is standard for microcontroller runtimes (TFLite-Micro~\citep{tflitemicro2021},
CMSIS-NN~\citep{cmsisnn2018}, and ESP-NN~\citep{espnn}) and for tiny models such as
MCUNet~\citep{mcunet2020,mcunetv2}. The fold is precisely what removes the statistics
\method restores.

\paragraph{Positioning.} Table~\ref{tab:position} summarizes the landscape against the
properties a method needs to actually adapt a \emph{deployed} MCU model. Each prior
method satisfies a strict subset; ours is the only one that is forward-only, runs on the
deployed int8-convolution model on a microcontroller, and reports measured energy.

\begin{table}[t]\centering\small
\caption{What it takes to adapt a deployed integer-only MCU model. ``int-only'' = runs on
the deployed int8-convolution MCU model (bit-exact integer convolutions, not GPU/simulated;
the forward-only adaptation itself may add fp32 as \method does); ``energy'' = measured
on-device energy. Each prior method satisfies only a subset.}
\label{tab:position}
\setlength{\tabcolsep}{3.5pt}
\begin{tabular}{lccccc}\toprule
 & fwd- & no & int- & MCU & meas. \\
method & only & BN & only & dep. & energy \\\midrule
TENT~\citep{tent2021}, CoTTA~\citep{cotta2022} & \xmark & \xmark & \xmark & \xmark & \xmark \\
BN-adapt~\citep{bnadapt2020} & \cmark & \xmark & \xmark & \xmark & \xmark \\
LeanTTA~\citep{leantta2025} & \cmark & \xmark & \xmark & \xmark & \xmark \\
FOA/ZOA/PACE~\citep{foa2024,zoa2025,pace2026} & \cmark & \cmark & \xmark & \xmark & \xmark \\
PEA~\citep{pea2026} & \cmark & \cmark & \xmark & \xmark & \xmark \\
TinyTTA~\citep{tinytta2024} & \xmark & \xmark & \xmark & \cmark & \xmark \\
\textbf{\method (ours)} & \cmark & \cmark & \cmark & \cmark & \cmark \\\bottomrule
\end{tabular}
\end{table}

\section{Method}
\label{sec:method}
\subsection{Folded integer-only deployment}
For efficient MCU inference, each \texttt{Conv$\to$BN} pair is fused into a single
convolution with bias: $W' = W\,\gamma/\sqrt{\sigma^2+\epsilon}$,
$b' = \beta - \mu\,\gamma/\sqrt{\sigma^2+\epsilon}$, and the BN layer is removed. This is
exact at inference time, so clean accuracy is unchanged.

\subsection{The adaptation gap}
\label{sec:gap}
\begin{figure}[t]\centering
\includegraphics[width=0.62\linewidth]{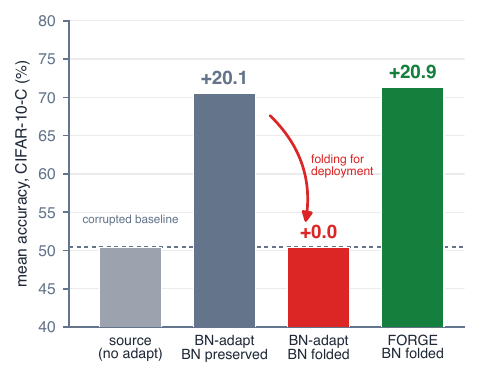}
\caption{\textbf{The adaptation gap.} On CIFAR-10-C (int8, mean over the 15 corruptions),
forward-only BN recalibration recovers $+20.1$ points on a BN-\emph{preserving} model, but
folding BN into the convolutions for integer deployment drops that to $+0.0$ (red): nothing
remains to recalibrate, and the BN-preserving model is itself not deployable in integer-only
form. \method restores $+20.9$ on the \emph{same} deployed folded model (green;
Sec.~\ref{sec:exp-baselines}). The corruption error is identical across all bars.}
\label{fig:gap}
\end{figure}

Folding eliminates the running statistics $(\mu,\sigma^2)$ that BN-based adaptation
updates. Empirically (Fig.~\ref{fig:gap}), forward-only BN recalibration recovers
$+20.1$ mean accuracy points on a BN-\emph{preserving} int8 model, but $+0.0$ once BN is
folded, since nothing remains to recalibrate, even though the corruption error is
identical in both cases. The need for adaptation is unchanged. Only the remedy has vanished.

\subsection{\method: per-channel recalibration}
We restore adaptation on the folded model by re-introducing a lightweight, gradient-free
per-channel correction where each BN used to be.

\paragraph{What folding removes, and what it leaves.} Before folding, a Conv$\to$BN site
normalizes its pre-activation to zero mean and unit variance using the running statistics
$(\mu,\sigma^2)$, then applies the affine $(\gamma,\beta)$; under distribution shift,
BN-adapt simply refreshes $(\mu,\sigma^2)$ from the test stream. Folding substitutes the
\emph{training-time} $(\mu,\sigma^2)$ into the convolution weights (Sec.~\ref{sec:method}.1)
and deletes the layer, so $(\mu,\sigma^2)$ are no longer addressable: there is nothing for
BN-adapt to refresh. What folding does \emph{not} change is the clean per-channel output
distribution of the fused convolution, which by construction still has mean $\beta_c$ and
standard deviation $|\gamma_c|$. \method keeps exactly those two per-channel constants,
recorded at fold time, as the recalibration target; no BN layer or stored feature bank is
needed.

\begin{figure}[t]
\centering
\includegraphics[width=0.62\linewidth]{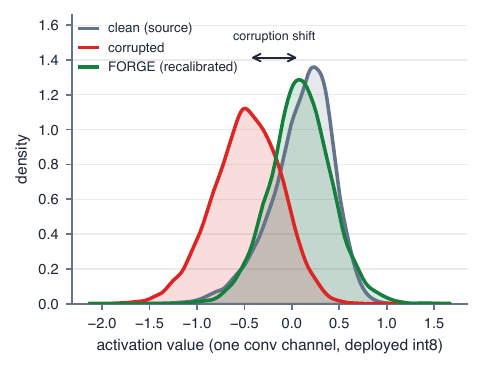}
\caption{\textbf{The mechanism on real activations.} Measured per-channel output distribution
of one convolution channel in the deployed int8 model (\texttt{layer2.0.bn1}, CIFAR-10-C
Gaussian noise). Corruption displaces the channel by $1.8\sigma$ off its clean source mean
(red); \method's forward-only recalibration restores it to within $0.06\sigma$ of the clean
distribution (green), recovering both the mean and the spread. This is the per-channel
correction of Alg.~\ref{alg:forge} measured on a real channel, not a schematic.}
\label{fig:act}
\end{figure}

\paragraph{Recalibration.} At test time we maintain an exponential moving average (EMA) of
the per-channel output mean and variance over the (shifted) stream and re-normalize each
channel back to its clean target (Alg.~\ref{alg:forge}):
\[
\hat{x}_c = \frac{x_c-\bar{\mu}_c}{\sqrt{\bar{\sigma}^2_c+\epsilon}}\,|\gamma_c| + \beta_c,
\]
where $(\bar\mu_c,\bar\sigma^2_c)$ are the running estimates, initialized to the clean
target so an unshifted stream is a no-op. On a real channel of the deployed int8 model this
snaps the corrupted output distribution back onto its clean target (Fig.~\ref{fig:act}).
This uses only forward passes and no learnable
parameters, and unlike BN recalibration it requires no BN layers, only the
$(\beta,|\gamma|)$ recorded at fold time. The cost is two reductions and an affine over the
activation, gradient-free and carrying no optimizer state. \method's recalibration runs in
fp32 on the MCU's FPU and is a small addition \emph{around} the integer convolution path,
not part of it. The deployed convolutions are genuine int8: int8 weights and activations
accumulated in int32 and requantized to int8 by integer arithmetic (ESP-NN), validated
bit-exact against an integer reference (Sec.~\ref{sec:device}). At each former-BN site the
int8 convolution output is dequantized to fp32, the running estimates
$(\bar\mu_c,\bar\sigma^2_c)$ and the affine of Alg.~\ref{alg:forge} are computed in fp32,
and the result is re-quantized to int8 for the next convolution. The only added state is
$2C$ fp32 scalars per site; there is no autograd graph, no master weights, and no optimizer
state, and this added fp32 cost is included in the measured on-device energy
(Sec.~\ref{sec:device}). We therefore describe \method precisely as forward-only adaptation
of an int8-convolution model \emph{using fp32 statistics}: ``integer-only'' refers to the
convolution engine \method runs inside, not to the arithmetic of the recalibration itself,
which is mixed-precision (int8 convolutions, fp32 recalibration). A naive alternative,
adapting the activation quantization scales (``scale-adapt'', a baseline we introduce
rather than a prior method, hence absent from the Table~\ref{tab:position} taxonomy),
fails (Table~\ref{tab:baselines}): a per-tensor scale only stretches the dynamic range and
cannot correct the per-channel mean/variance shift that corruption induces.

\begin{algorithm}[t]
\caption{\method recalibration at one former-BN site}
\label{alg:forge}
\begin{algorithmic}[1]
\Require folded conv output $x\!\in\!\mathbb{R}^{B\times C\times H\times W}$; clean targets
$(\beta_c,|\gamma_c|)$ from fold time; momentum $m$; running $(\bar\mu_c,\bar\sigma^2_c)$
(init.\ to $(\beta_c,\gamma_c^2)$)
\State $\mu_c \gets \mathrm{mean}_{B,H,W}(x_c)$;\quad $v_c \gets \mathrm{var}_{B,H,W}(x_c)$
\Comment{per-channel batch stats}
\State $\bar\mu_c \gets (1{-}m)\,\bar\mu_c + m\,\mu_c$;\quad
$\bar\sigma^2_c \gets (1{-}m)\,\bar\sigma^2_c + m\,v_c$ \Comment{forward-only EMA}
\State $\hat{x}_c \gets \dfrac{x_c-\bar\mu_c}{\sqrt{\bar\sigma^2_c+\epsilon}}\,|\gamma_c| + \beta_c$
\Comment{re-center to clean target}
\State \Return $\hat{x}_c$ \quad(no gradients, no learnable parameters)
\end{algorithmic}
\end{algorithm}

\paragraph{Selective-layer recalibration.}
\label{sec:selective}
Per-layer importance is highly non-uniform: a single layer's recovery ranges from
$+0.4$ to $+17$ points. Adapting only the most important layers is both cheaper and,
because the unhelpful layers slightly hurt, \emph{more accurate} than adapting all
layers (Sec.~\ref{sec:exp-selective}).

\paragraph{Single-sample streaming.}
The deployment regime is batch size $1$. Single-image statistics are noisy, so a fixed
EMA momentum collapses; the effective averaging window is $\approx
\text{batch}/\text{momentum}$, so we scale momentum with batch size to hold the window
constant (Sec.~\ref{sec:exp-bs}).

\section{Experiments}
\label{sec:exp}
\paragraph{Setup.} ResNet-20~\citep{resnet2016} and MobileNetV2~\citep{mobilenetv2,mobilenetv3}
($0.5\times$) on CIFAR-10/100~\citep{cifar2009} and Tiny-ImageNet (a 200-class,
$64{\times}64$ ImageNet~\citep{imagenet2009} subset); corruption benchmarks CIFAR-10/100-C
and Tiny-ImageNet-C~\citep{cifarc2019} at severity 5. Models are folded
and quantized to int8 (per-channel weights, per-tensor activations). We report mean
accuracy / recovery over the 15 corruptions. Because online adaptation is
order-dependent, recovery is reported as mean$\,\pm\,$std over 5 random test-stream
orderings (Tables~\ref{tab:baselines},~\ref{tab:general}). On-device: ESP32-S3, energy
via a Nordic PPK2. Some ablations use a fixed subset for compute; full-set runs are noted.

\subsection{Comparison to baselines}
\label{sec:exp-baselines}
\method matches forward-only BN adaptation and recovers most of backprop-based TENT's
gain ($+20.9$ vs.\ $+24.9$), while being the only method that runs on the deployed folded
integer-only model (Table~\ref{tab:baselines}). The $\sim$4-point gap to TENT is the
price of being gradient-free: TENT takes a gradient step per sample and keeps improving
over the stream, whereas our running-statistic estimate converges and plateaus. This
gap is, however, the wrong comparison for the deployment setting: neither TENT (it needs
gradients) nor BN-adapt (it needs BN layers) can run on the folded integer-only model at
all. On the device, the practical alternative to our $+20.9$ is not TENT but the
unadapted model ($+0.0$; the folded source of Fig.~\ref{fig:gap}). We also evaluate
CoTTA~\citep{cotta2022} in its native continual protocol (the 15 corruptions as one
stream): it recovers $+13.9$ points, but like TENT it relies on backpropagation
(and augmentation forward passes), so it too cannot run on the folded integer-only model
(Table~\ref{tab:position}).

\begin{table}[t]\centering\small
\caption{Baselines on CIFAR-10-C (bs=64, full test set). The top group adapts the fp32
model (recovery vs the fp32 source); the bottom group adapts the deployed folded int8
model (recovery vs its int8 source, $50.4\%$). Only \method is forward-only, needs no BN
layers, \emph{and} runs on the folded integer-only MCU model. Subscripts are std over 5
random test-stream orderings. This table reports \emph{measured accuracy}, so it also
includes the unadapted source models and \emph{scale-adapt} (a naive int8 baseline that
adapts the per-tensor activation quantization scales, introduced here in
Sec.~\ref{sec:method}); these are not adaptation methods from the literature and so do not
appear in the Table~\ref{tab:position} capability taxonomy, which lists only prior TTA
methods and ours.}
\label{tab:baselines}
\setlength{\tabcolsep}{3.5pt}
\begin{tabular}{lccccc}\toprule
method & acc.\,(\%) & rec. & fwd & no BN & MCU \\\midrule
source (fp32)    & 50.1 & ---     & \cmark & \cmark & \cmark \\
BN-adapt         & 70.7 & $+20.6_{\pm 0.09}$ & \cmark & \xmark & \xmark \\
TENT (backprop)  & 75.0 & $+24.9_{\pm 0.13}$ & \xmark & \xmark & \xmark \\\midrule
source (int8)    & 50.4 & ---     & \cmark & \cmark & \cmark \\
scale-adapt      & 50.0 & $-0.4$  & \cmark & \cmark & \cmark \\
\textbf{\method} & \textbf{71.3} & $+20.9_{\pm 0.04}$ & \cmark & \cmark & \cmark \\\bottomrule
\end{tabular}
\end{table}

\paragraph{Per-corruption behaviour.} Table~\ref{tab:percorr} breaks the comparison out
over all 15 corruptions. \method tracks forward-only BN-adapt closely on every corruption
while running on the folded int8 model, and the gains are largest exactly where they
matter most: severe corruptions that collapse the source model (contrast $25.4\!\to\!79.0$,
Gaussian noise $25.6\!\to\!60.3$). On already-mild corruptions where the source is strong
(brightness, JPEG) it is flat to slightly negative, the standard behaviour of
normalization-based adaptation, which the safety gate of Sec.~\ref{sec:gate} removes.

\begin{table}[t]\centering\small
\setlength{\tabcolsep}{4.5pt}
\caption{Per-corruption accuracy (\%) on CIFAR-10-C (severity 5, bs=64). ``source'' is the
fp32 model; \method runs on the folded int8 model. \method matches forward-only BN-adapt
across the board and recovers most on the hardest corruptions.}
\label{tab:percorr}
\begin{tabular}{lcccc}\toprule
corruption & source & BN-adapt & TENT & \method \\\midrule
Gaussian noise & 25.6 & 59.7 & 67.3 & 60.3 \\
Shot noise     & 31.1 & 60.6 & 69.3 & 61.4 \\
Impulse noise  & 30.8 & 54.0 & 61.1 & 54.3 \\
Defocus blur   & 45.0 & 81.4 & 83.2 & 82.0 \\
Glass blur     & 40.6 & 56.5 & 60.5 & 56.6 \\
Motion blur    & 56.1 & 78.7 & 81.1 & 79.4 \\
Zoom blur      & 49.5 & 80.3 & 83.2 & 80.6 \\
Snow           & 67.5 & 71.7 & 76.8 & 72.4 \\
Frost          & 52.1 & 71.7 & 76.3 & 72.4 \\
Fog            & 63.4 & 77.4 & 80.4 & 78.1 \\
Brightness     & 86.8 & 84.8 & 87.1 & 85.7 \\
Contrast       & 25.4 & 78.6 & 79.2 & 79.0 \\
Elastic        & 67.4 & 68.2 & 71.7 & 68.9 \\
Pixelate       & 41.4 & 70.2 & 76.7 & 70.7 \\
JPEG           & 69.0 & 66.2 & 71.3 & 67.2 \\\midrule
mean           & 50.1 & 70.7 & 75.0 & 71.3 \\\bottomrule
\end{tabular}
\end{table}

\subsection{Selective-layer recalibration}
\label{sec:exp-selective}
\begin{figure}[t]\centering
\includegraphics[width=0.62\linewidth]{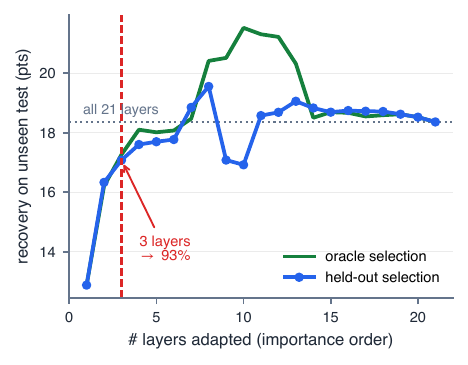}
\caption{Held-out selective recalibration (full CIFAR-10-C test). Layers ranked on
held-out corruptions; recovery shown on \emph{unseen} test corruptions. A few layers
reach $93\%$ of full recovery (knee at 3), held-out tracks the oracle, and the best
subset exceeds adapting all 21 layers.}
\label{fig:selective}
\end{figure}

We stress that layer selection is a \emph{one-time, design-time} step performed by the
developer \emph{before} deployment, using only source data and held-out synthetic
corruptions: it never touches the test stream and never runs on the device, so it does not
require test-time data and does not conflict with the deployed-model, no-data premise of
TTA. It is also \emph{optional}: adapting \emph{all} 21 layers needs no selection or
calibration set whatsoever and already recovers $+18.4$ (below); selection is a cheap
offline optimization on top, not a prerequisite. Concretely, ranking layer importance on a
held-out set of corruptions and evaluating on a disjoint set, five of the six most important
layers are shared across the split (importance is largely corruption-independent, so the
ranking transfers without ever seeing the test corruptions). Adapting the top-3 layers
(chosen on held-out corruptions) recovers $93\%$ of the full benefit on unseen corruptions,
matching an oracle that sees the test set (Fig.~\ref{fig:selective}).
Adapting the best subset \emph{exceeds} adapting all 21 layers (the held-out selection
peaks at $+19.6$ with 8 layers versus $+18.4$ for all 21), because the unhelpful layers
slightly hurt.

\subsection{Single-sample streaming}
\label{sec:exp-bs}
\begin{figure}[t]\centering
\includegraphics[width=0.62\linewidth]{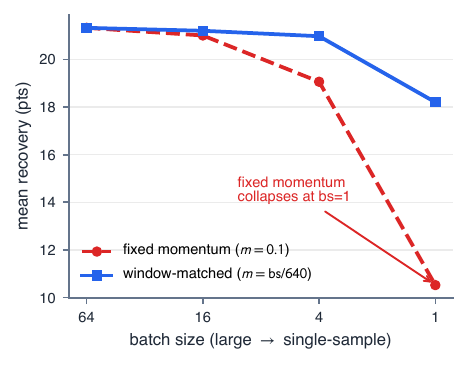}
\caption{Recovery vs.\ batch size (CIFAR-10-C). A fixed EMA momentum collapses at
single-sample streaming; scaling momentum with batch size (constant averaging window
$\approx\text{bs}/m$) holds.}
\label{fig:bs}
\end{figure}

A fixed momentum holds down to batch 4 but collapses at batch 1; window-matched momentum
($m=\text{bs}/640$) holds (Fig.~\ref{fig:bs}).

\subsection{Generalization}
\method generalizes across datasets and architectures (Table~\ref{tab:general}):
ResNet-20 and MobileNetV2 (a depthwise-separable, inverted-residual family) recover
nearly identically on CIFAR-10-C, and the method holds on the harder 100-class
CIFAR-100-C. The MobileNetV2 result matters beyond a second data point: the recalibration
does \emph{not} rely on standard (``vanilla'') convolutions---it applies unchanged to the
depthwise and pointwise layers that make up the efficient, separable backbones used on edge
devices, since every such layer is still a folded \texttt{Conv}$\to$\texttt{BN} site with a
per-channel output distribution to restore. It also scales to Tiny-ImageNet-C (200 classes, $64{\times}64$; corruptions
generated with the \texttt{imagecorruptions} library~\citep{imagecorruptions2019}), where it
recovers $+11.0$ points with \emph{all} 15 corruptions improving. This is a smaller
absolute gain, as expected on a much harder benchmark ($53\%$ clean) where severity-5
corruptions are devastating, but it is a clear positive across the board.

\begin{table}[t]\centering\small
\caption{Generalization: mean recovery (bs=64). Two architectures $\times$ three datasets
(CIFAR-10-C, CIFAR-100-C, and Tiny-ImageNet-C / TIN-C: 200 classes, $64{\times}64$).
Subscripts are std over 5 test-stream orderings; TIN-C is single-seed.}
\label{tab:general}
\begin{tabular}{lccc}\toprule
 & CIFAR-10-C & CIFAR-100-C & TIN-C \\\midrule
ResNet-20 & $+20.9_{\pm 0.04}$ & $+14.5_{\pm 0.06}$ & --- \\
MobileNetV2 & $+20.5_{\pm 0.09}$ & --- & $+11.0$ \\\bottomrule
\end{tabular}
\end{table}

\subsection{Ablations}
\label{sec:ablations}
\paragraph{Momentum.} \method has a single hyperparameter, the EMA momentum. Recovery is
remarkably flat in it (Table~\ref{tab:abl-mom}): sweeping $m$ across two orders of magnitude
($0.01$ to $1.0$) moves mean recovery by under $0.8$ points, peaking at $m{=}0.1$. The method
therefore needs no per-deployment tuning of its only knob.

\paragraph{Bit width.} Recalibration is not specific to int8. As weight/activation
precision drops (Table~\ref{tab:abl-bits}), \method keeps recovering: at aggressive 4-bit
weights it recovers $+23.3$ (the quantized source is lower, so there is more to restore),
and even at 4/4---where quantization alone collapses clean accuracy to $60.6\%$---it still
recovers $+13.3$. The recovery magnitude tracks the available headroom, but the mechanism
itself never breaks, which is what we would expect of a correction that re-centers
distributions rather than relying on a particular numeric range.

\begin{table}[t]\centering\small
\setlength{\tabcolsep}{4.5pt}
\caption{Momentum ablation (ResNet-20, CIFAR-10-C, int8, bs=64). Mean recovery vs.\ the int8
source is stable across two orders of magnitude of the EMA momentum.}
\label{tab:abl-mom}
\begin{tabular}{lccccc}\toprule
momentum $m$ & 0.01 & 0.03 & 0.1 & 0.3 & 1.0 \\\midrule
recovery & $+20.2$ & $+20.4$ & $+20.8$ & $+20.7$ & $+20.1$ \\\bottomrule
\end{tabular}
\end{table}

\begin{table}[t]\centering\small
\setlength{\tabcolsep}{6pt}
\caption{Bit-width ablation (ResNet-20, CIFAR-10-C). \method recovers across precisions;
the magnitude tracks the headroom (lower source $=$ more to restore) and the mechanism holds
even at 4-bit weights and activations.}
\label{tab:abl-bits}
\begin{tabular}{lcccc}\toprule
(weight, act.)\ bits & 8/8 & 6/6 & 4/8 & 4/4 \\\midrule
clean acc.\ (\%)  & 91.9 & 90.5 & 90.1 & 60.6 \\
source acc.\ (\%) & 50.4 & 50.2 & 47.6 & 32.2 \\
recovery          & $+20.8$ & $+18.5$ & $+23.3$ & $+13.3$ \\\bottomrule
\end{tabular}
\end{table}

\subsection{When to adapt: a safety gate}
\label{sec:gate}
Normalization-based adaptation can slightly hurt on already-mild corruptions where the
deployed model is fine (Table~\ref{tab:percorr}: brightness, JPEG). A forward-only gate
removes this downside. The naive signal, absolute prediction confidence, does not separate
the cases: JPEG (mean top-1 confidence $0.855$, adaptation hurts) and snow ($0.857$,
adaptation helps $+4.8$) are indistinguishable, so a confidence threshold either fails to
protect JPEG or needlessly drops snow. The signal that does work is whether adaptation
\emph{improves} the model's own confidence. Obtaining the adapted and unadapted confidence
does require both forward passes, so we do this only during a short calibration window of
$W$ batches at the start of a stream: we run both paths, accumulate each one's mean top-1
confidence, and if adaptation raises mean confidence by a margin $\delta$ we commit to it
for the rest of the stream, otherwise we revert to the source (Alg.~\ref{alg:gate}). The
gate is thus causal and online: the two paths run together only inside the window (a
$2{\times}$ forward cost over $W$ batches, single-path afterward), it needs no labels, and
its only added state is two scalar confidence accumulators. On CIFAR-10-C it keeps
adaptation on $12$ of $15$ streams and reverts on the rest (including brightness and JPEG,
the two where adaptation would hurt), turning the 2 degraded corruptions into 0 while
slightly \emph{raising} mean recovery ($+20.8\!\to\!+21.0$; Table~\ref{tab:gate}).

We evaluated the gate across all three benchmarks (Table~\ref{tab:gate}). It degrades
\emph{no} corruption on any of them, but its safety is not free, and the
confidence-improvement signal weakens as the number of classes grows: it keeps adaptation
on $12/15$ streams on CIFAR-10-C, $5/15$ on CIFAR-100-C, and $0/15$ on Tiny-ImageNet-C. On
CIFAR-100-C the gate trades recovery for safety ($+14.4$ ungated to $+5.9$ gated). On
Tiny-ImageNet-C the deployed MobileNetV2 is badly \emph{overconfident} on its errors, so
adaptation \emph{raises accuracy but lowers mean confidence on every corruption}; the gate
therefore never fires and forfeits the full $+11.0$ recovery (Table~\ref{tab:general}) while
still degrading nothing. We therefore claim the gate is safe \emph{under the conditions we
evaluate}, not unconditionally, and useful mainly as a conservative safeguard; a
self-supervised signal that tracks accuracy in the many-class overconfident regime, and an
on-device latency/energy characterization of the window, are future work.

\begin{table}[t]\centering\small
\setlength{\tabcolsep}{5pt}
\caption{When-to-adapt gate across the three benchmarks (int8; ResNet-20 for CIFAR,
MobileNetV2 for Tiny-ImageNet; confidence-improvement rule, margin $\delta{=}0$). The gate
degrades no corruption on any benchmark, but the confidence signal weakens with class count:
it adapts fewer streams and, on Tiny-ImageNet-C, none. ``fires'' = streams kept adapted;
Tiny-ImageNet-C ungated recovery is from Table~\ref{tab:general} (the overconfident source
makes the gate revert everywhere, so gated recovery is $0$). Single representative run per
benchmark, so the CIFAR-10-C $+20.8$ is within seed noise of the $+20.9$ in
Table~\ref{tab:baselines}.}
\label{tab:gate}
\begin{tabular}{lccc}\toprule
benchmark & ungated rec.\ (hurt) & fires & gated rec.\ (hurt) \\\midrule
CIFAR-10-C       & $+20.8$ (2/15) & 12/15 & $\mathbf{+21.0}$ (0/15) \\
CIFAR-100-C      & $+14.4$ (1/15) &  5/15 & $+5.9$ (0/15) \\
Tiny-ImageNet-C  & $+11.0$ (0/15) &  0/15 & $+0.0$ (0/15) \\\bottomrule
\end{tabular}
\end{table}

\begin{algorithm}[t]
\caption{Causal when-to-adapt gate (per stream)}
\label{alg:gate}
\begin{algorithmic}[1]
\Require calibration window $W$ (batches); margin $\delta$; recalib model $f_{\text{rc}}$ (updates EMA), source $f$
\State $s_{\text{on}}\gets0;\ s_{\text{off}}\gets0;\ n\gets0$ \Comment{two scalar confidence accumulators}
\For{the first $W$ batches $x_t$ of the stream}
  \State $s_{\text{off}}\mathrel{+}= \overline{\max\,\mathrm{softmax}\,f(x_t)}$ \Comment{unadapted forward}
  \State $s_{\text{on}}\mathrel{+}= \overline{\max\,\mathrm{softmax}\,f_{\text{rc}}(x_t)}$;\quad $n\mathrel{+}=1$ \Comment{adapted forward (updates EMA)}
\EndFor
\State $\textsc{adapt}\gets [\,(s_{\text{on}}-s_{\text{off}})/n > \delta\,]$ \Comment{commit for the rest of the stream}
\For{each remaining batch $x_t$}
  \State \Return $f_{\text{rc}}(x_t)$ \textbf{if} \textsc{adapt} \textbf{else} $f(x_t)$ \Comment{single path, no extra cost}
\EndFor
\end{algorithmic}
\end{algorithm}

\subsection{On-device deployment}
\label{sec:device}
We export the folded int8 model and verify an integer-convolution reference (int8 weights
and activations, int32 accumulation, integer requantization) reproduces the
PyTorch path (clean $92.6\%$ vs.\ $92.2\%$). On the ESP32-S3 the C inference matches the
reference logits exactly ($\max|\Delta|=0.0$), so the deployed model is the same one the
simulation results describe. Inference uses the ESP-NN SIMD-optimized int8 kernels, and the recalibration itself
\emph{runs on the physical ESP32-S3, not in a simulator}: both the inference and the
inference+adapt numbers below are measured on the same on-device firmware whose logits match
the integer reference exactly. Energy is measured with a Nordic PPK2 in source mode
(3.3\,V), integrating current at $100$\,kHz; inference and inference+adapt are separated by
their distinct durations.

Inference runs on the optimized ESP-NN SIMD int8 kernels at $286$\,ms ($9.6\times$ faster
than a naive convolution), and adaptation adds only $8.3$\,mJ ($6.8\%$ of inference energy)
and $21.9$\,ms (Table~\ref{tab:energy}): forward-only adaptation is cheap on a
microcontroller. The reason is structural. \method adds, per former-BN site, two
per-channel reductions (mean and variance) and one per-channel affine over the activation;
there is no weight update, no second forward pass, and no gradient buffer, so the overhead
is a \emph{fixed} $8.3$\,mJ regardless of how fast the convolutions run, scaling with the
\emph{activation} volume the network already streams through rather than the parameter
count. Memory is equally modest, and we account for it explicitly (measured from the
firmware). \method adds two kinds of state per channel: a \emph{mutable} running mean and
variance (SRAM), and the \emph{immutable} clean target $(\beta_c,|\gamma_c|)$ recorded at
fold time (Flash). Across all $21$ ResNet-20 sites ($784$ channels) that is $2{\times}784$
fp32 mutable SRAM ($6.1$\,KB) plus $2{\times}784$ fp32 immutable Flash ($6.1$\,KB); the
selective top-3 ($112$ channels) needs $0.9$\,KB each (breakdown below). Recalibration is
applied \emph{layer-by-layer and in place}: each site corrects its own activation as the
forward pass reaches it, reusing the fp32 activation buffer the firmware already holds (three
buffers of $16{\times}32{\times}32$, $192$\,KB, which dominate peak SRAM), so \method adds no
extra activation-sized scratch and does not raise peak memory. The headline contrast is not with TENT's accuracy but with its deployability:
backpropagation needs the gradient graph and optimizer state that an integer-only MCU
runtime cannot hold, whereas \method runs in the same forward pass as inference, adding
only a lightweight fp32 recalibration around the int8 convolutions. No prior forward-only
TTA method reports this measured on-device cost.

\begin{table}[!ht]\centering\small
\caption{Measured on-device energy (ESP32-S3, PPK2, 3.3\,V) with the deployed ESP-NN SIMD
int8 kernels. The adaptation overhead is a fixed cost, independent of the inference kernel.}
\label{tab:energy}
\begin{tabular}{lccc}\toprule
operation & energy (mJ) & time (ms) & power (mW) \\\midrule
inference & 121.7 & 286.1 & 425 \\
inference + adapt & 130.0 & 308.0 & 422 \\\midrule
\textbf{adapt overhead} & \textbf{8.3} & \textbf{21.9} & --- \\\bottomrule
\end{tabular}
\end{table}

\par\medskip
\noindent\begin{minipage}{\linewidth}
\centering
\footnotesize
\textbf{Measured memory footprint of \method on the ESP32-S3 firmware} (fp32, 4\,B/value).
The activation buffers are held by inference regardless of adaptation; \method adds only the
running-stat SRAM and the target Flash, and no activation-sized scratch (in-place).\\[3pt]
\setlength{\tabcolsep}{6pt}
\begin{tabular}{lccc}\toprule
component & data type & all 21 sites (784 ch) & top-3 (112 ch) \\\midrule
mutable running mean+var (SRAM) & fp32 & $6.1$\,KB & $0.9$\,KB \\
immutable targets $(\beta,|\gamma|)$ (Flash) & fp32 & $6.1$\,KB & $0.9$\,KB \\
recalibration scratch & --- & none (in place) & none (in place) \\
activation buffers (peak, inference) & fp32 & $192$\,KB & $192$\,KB \\\midrule
\textbf{added by \method} & & \textbf{12.2\,KB} & \textbf{1.8\,KB} \\\bottomrule
\end{tabular}
\end{minipage}\par\medskip

\section{Limitations}
\label{sec:limits}
The recalibration mechanism is close to BN-statistic adaptation; our contribution is
making it work on a \emph{deployed} folded integer-only model and measuring its
on-device cost, not a new adaptation principle. \method is \emph{mixed-precision by design}:
the convolutions are int8 (validated bit-exact on the ESP32-S3), but the recalibration
itself runs in fp32 on the FPU, so we claim integer-only \emph{convolution} execution, not
an end-to-end integer-only adaptation path. The confidence gate of Sec.~\ref{sec:gate}
degrades no corruption on any of the three benchmarks, but its confidence-improvement signal
weakens as class count grows: it trades recovery for safety on CIFAR-100-C and does not fire
at all on Tiny-ImageNet-C, where the overconfident source decouples confidence from accuracy.
A signal that tracks accuracy in that regime, and an on-device latency/energy
characterization of the calibration window, are future work. On-device measurement is
on ResNet-20 while the second architecture is evaluated in simulation. \method also targets the BN-fold regime specifically: it assumes
folded \texttt{Conv}$\to$\texttt{BN} sites, so it covers the convolutional backbones common
on microcontrollers (including the depthwise-separable family above) but not transformer
backbones, whose LayerNorm is not fused into a preceding convolution in the same way.
Extending the same post-fold restoration idea to other normalization layers is natural
future work.

\section{Conclusion}
Deploying a vision model on a microcontroller, which folds BN and quantizes to
integers, silently removes its ability to adapt to the distribution shift it will face
in the field. We restore that ability with a forward-only per-channel recalibration that
runs on the deployed int8-convolution model (a lightweight fp32 correction around the
integer convolutions), matches gradient-based adaptation in accuracy,
needs only a few layers, survives single-sample streaming, generalizes across datasets
and architectures, and, measured on real hardware, costs $6.8\%$ of inference energy.
Forward-only test-time adaptation is practical, and essentially free, on a microcontroller.

\appendix
\section{Why recalibration cancels the corruption, and the variance it omits}
\label{app:derivation}

\paragraph{Setup.} At a folded \texttt{Conv}$\to$\texttt{BN} site, the clean output of
channel $c$ is $y_c = \gamma_c z_c + \beta_c$, where $z_c=(u_c-\mu_c)/\sqrt{\sigma^2_c+\epsilon}$
is the batch-normalized pre-activation. By construction $z_c$ has zero mean and unit variance
on the training distribution, so $\mathbb{E}[y_c]=\beta_c$ and $\mathrm{Std}[y_c]=|\gamma_c|$:
these are the two per-channel constants \method records at fold time, and the only quantities
it retains about the clean model.

\paragraph{Exact cancellation under an affine channel shift.} To first order, a covariate
shift propagates through one folded linear$+$ReLU site as a per-channel affine map of the
clean output, $\tilde{y}_c = a_c\,y_c + d_c$ with scale $a_c>0$ and offset $d_c$---exactly the
``$x = A\,x_{\text{orig}} + B$'' regime. Its moments are
$\mathbb{E}[\tilde{y}_c]=a_c\beta_c+d_c$ and $\mathrm{Var}[\tilde{y}_c]=a_c^2\gamma_c^2$.
\method's running estimates converge to these, $\bar\mu_c\!\to\!\mathbb{E}[\tilde{y}_c]$ and
$\bar\sigma^2_c\!\to\!\mathrm{Var}[\tilde{y}_c]$, and the recalibration (Alg.~\ref{alg:forge})
gives, as $\epsilon\!\to\!0$,
\[
\hat{x}_c=\frac{\tilde{y}_c-\bar\mu_c}{\sqrt{\bar\sigma^2_c+\epsilon}}\,|\gamma_c|+\beta_c
=\frac{a_c y_c+d_c-(a_c\beta_c+d_c)}{\sqrt{a_c^2\gamma_c^2}}\,|\gamma_c|+\beta_c
=\frac{a_c\,(y_c-\beta_c)}{a_c|\gamma_c|}\,|\gamma_c|+\beta_c=y_c .
\]
The clean output is recovered \emph{exactly}, and the corruption coefficients cancel: the
standardization divides out the scale $a_c$ independently of its value, and subtracting
$\bar\mu_c$ removes the offset $d_c$. Tracking $(\bar\mu_c,\bar\sigma^2_c)$ and rescaling thus
acts as the inverse of the affine corruption map, dropping the multiplicative coefficient---no
property of the corruption need be known, only the fold-time $(\beta_c,|\gamma_c|)$.

\paragraph{The variance the EMA omits (law of total variance).} The cancellation above uses
the \emph{true} $\mathrm{Var}[\tilde{y}_c]$. \method estimates it with an EMA of per-batch
variances $v_i=\mathrm{Var}(\tilde{y}_c\mid\text{batch }i)$, which for a stationary stream
converges to $\mathbb{E}[\mathrm{Var}(\tilde{y}_c\mid\text{Batch})]$. By the law of total
variance,
\[
\mathrm{Var}(\tilde{y}_c)=\underbrace{\mathbb{E}[\mathrm{Var}(\tilde{y}_c\mid\text{Batch})]}_{\text{tracked by the EMA}}
+\underbrace{\mathrm{Var}(\mathbb{E}[\tilde{y}_c\mid\text{Batch}])}_{\text{omitted}},
\]
so the estimate falls short by the between-batch-mean term, and the recalibration divides by a
slightly \emph{under}-estimated standard deviation. If the omitted term is a fraction $f$ of
the total variance, the residual scaling error is $1/\sqrt{1-f}-1\approx f/2$. For random
batches of size $B$ from a stream whose per-image channel means have variance $\sigma^2_\mu$,
the omitted term is exactly $\mathrm{Var}(\mathbb{E}[\tilde{y}_c\mid\text{Batch}])
=(\sigma^2_\mu/B)\,(N-B)/(N-1)$: it shrinks as $1/B$.

We measure $f$ directly on the deployed int8 model, over $783$ of the $784$ channels of the
$21$ recalibration sites (one channel has near-constant output and is excluded;
Table~\ref{tab:lotv}). At the batch size of our main results ($B{=}64$)
the omitted term is $\approx\!0.1\%$ of the total variance---a $\approx\!0.05\%$ scaling
error, far below the int8 quantization step itself---so the cancellation holds to high
precision. As $B$ falls the term grows as $1/B$, reaching $\approx\!8\%$ (a $\approx\!4.7\%$
scaling error) at single-sample streaming on Gaussian noise, and more for the most severe
shifts ($16\%$ for contrast, $10\%$ for fog). This is precisely why a fixed-momentum estimator
degrades at $B{=}1$ (Sec.~\ref{sec:exp-bs}, Fig.~\ref{fig:bs}): the window-matched momentum
$m=\text{bs}/640$ we adopt holds the effective averaging window---and hence this between-batch
term---constant as the batch shrinks, which is the practical control for the bias the
decomposition identifies.

\begin{table}[H]\centering\small
\caption{The between-batch variance term the EMA omits, measured on the deployed int8 model
(ResNet-20, CIFAR-10-C Gaussian noise, severity 5; mean over $783$ of the $784$ channels of
the $21$ recalibration sites, excluding one near-constant channel). The omitted term is a fraction of the true total variance and shrinks as
$1/B$; the implied error in the $1/\mathrm{std}$ rescale is $\approx\!f/2$. It is negligible at
the $B{=}64$ of the main results and dominates only at single-sample streaming, where the
window-matched momentum of Sec.~\ref{sec:exp-bs} controls it.}
\label{tab:lotv}
\setlength{\tabcolsep}{5.5pt}
\begin{tabular}{lccccccc}\toprule
batch size $B$ & 1 & 2 & 4 & 8 & 16 & 32 & 64 \\\midrule
omitted $\mathrm{Var}(\mathbb{E}[X|B])/\mathrm{Var}(X)$ & $7.96\%$ & $3.97\%$ & $1.98\%$ & $0.98\%$ & $0.48\%$ & $0.23\%$ & $0.11\%$ \\
implied $1/\mathrm{std}$ scaling error & $4.73\%$ & $2.14\%$ & $1.03\%$ & $0.50\%$ & $0.24\%$ & $0.12\%$ & $0.05\%$ \\\bottomrule
\end{tabular}
\end{table}

{\small\bibliographystyle{tmlr}\bibliography{refs}}
\end{document}